\PassOptionsToPackage{numbers,sort&compress}{natbib}
\documentclass{article}

\usepackage[dblblindworkshop,final]{neurips_2026}
\usepackage[utf8]{inputenc} % allow utf-8 input
\usepackage[T1]{fontenc}    % use 8-bit T1 fonts
\usepackage{hyperref}       % hyperlinks
\usepackage{url}            % simple URL typesetting
\usepackage{booktabs}       % professional-quality tables
\usepackage{amsfonts}       % blackboard math symbols
\usepackage{nicefrac}       % compact symbols for 1/2, etc.
\usepackage{microtype}      % microtypography
\usepackage{xcolor}         % colors
\usepackage{soul}
\usepackage{amsmath}
\usepackage{graphicx}
\usepackage{siunitx}
\usepackage{multirow}
\usepackage{caption}
\usepackage{makecell}

\usepackage[final,inline,nomargin]{fixme}
\fxsetup{theme=color,mode=multiuser}
\FXRegisterAuthor{va}{vivi}{\color{blue}VA}

\newcommand{\pco}{p\mathrm{CO}_2}
\newcommand{\uatm}{\,\si{\micro atm}}
\newcommand{\best}[1]{\textbf{#1}}

\workshoptitle{Representations for the Physical Sciences}
\title{Learning Calibrated and Samplable Predictive Distributions of Surface-Ocean $\Delta p\mathrm{CO}_2$ Using Tabular Diffusion Models}
\author{
    Gabriele Accarino\textsuperscript{1}\thanks{Corresponding author: \texttt{ga2673@columbia.edu}} \quad
    Mohammad M. Amirian\textsuperscript{2} \quad
    Thea H. Heimdal\textsuperscript{3} \quad
    Amanda R. Fay\textsuperscript{3} \\
    \textbf{Galen A. McKinley\textsuperscript{3}} \quad
    \textbf{Viviana Acquaviva\textsuperscript{2,3}} \\
    [0.8em]
    \small\textsuperscript{1}Department of Earth and Environmental Engineering, Columbia University, New York, NY, USA \\
    \small\textsuperscript{2}CUNY New York City College of Technology, Brooklyn, NY, USA \\
    \small\textsuperscript{3}Lamont-Doherty Earth Observatory, Columbia University, Palisades, NY, USA\\
}
\begin{document}

\maketitle

\begin{abstract}
The ocean absorbs almost $30\%$ of anthropogenic $\mathrm{CO}_2$ emissions, but quantifying this sink requires estimating surface-ocean $\mathrm{CO}_2$ from sparse observations in space and time. Most existing machine-learning estimates of this field are deterministic, giving a single value per grid point, with uncertainty approximated post hoc via residuals, ensembles, or prescribed distributions. We instead learn, for each observation, a conditional predictive distribution: using Treeffuser, a conditional diffusion model with gradient-boosted-tree score functions, we estimate the full predictive distribution of $\Delta p\mathrm{CO}_2$ (the sea--air $p\mathrm{CO}_2$ difference) from $13$ environmental, spatial, and temporal predictors. On a held-out SOCAT test set, Treeffuser attains, among the uncertainty-quantification methods evaluated, the lowest continuous ranked probability score ($8.229$ \si{\micro atm}) and the coverage closest to the nominal $68\%$ / $95\%$ levels ($0.686$ / $0.930$), while retaining point-prediction skill comparable to a deterministic XGBoost baseline (RMSE $18.4$ vs $17.7$ \si{\micro atm}; $R^2$ $0.77$ vs $0.79$). These results show that conditional diffusion can provide well-calibrated, observation-level uncertainties while largely preserving point-prediction skill. Because Treeffuser is generative, its samples can support probabilistic analyses of $\Delta p\mathrm{CO}_2$; extending this to spatially coherent field reconstruction will require modeling dependence across grid cells and evaluation on spatially held-out data.

\end{abstract}
\section{Introduction}

Anthropogenic carbon dioxide (CO$_2$) emissions are the primary drivers of climate change. The ocean mitigates such effect by absorbing a large fraction of these emissions through physical \cite{wanninkhof2014relationship} and chemical processes \cite{zeebe2001co2}, while biological processes also play an important role in the ocean’s natural carbon cycle \cite{m2025parameterization, devred2025net}. However, substantial uncertainties remain in quantifying ocean carbon uptake \citep{mckinley2026global, friedlingstein2025global, McKinley2017}. Quantifying the ocean carbon sink requires resolving the air--sea CO$_2$ flux, whose direction and magnitude depend strongly on the difference between surface-ocean and atmospheric CO$_2$ partial pressures ($\Delta p\mathrm{CO}_2$), together with gas-transfer velocity and CO$_2$ solubility.

Decades of coordinated international observations have recorded millions of direct surface-ocean $p\mathrm{CO}_2$ measurements \citep{bakker2016socat}, but these sample only a small and uneven fraction of the ocean, requiring the reconstruction of the full spatio-temporal field. This problem can be formulated as supervised learning from heterogeneous environmental, spatial, and temporal predictors. Tree-based methods are particularly effective in this setting, as they capture nonlinear interactions with limited preprocessing. Prominent examples include the physically informed $p\mathrm{CO}_2$-Residual XGBoost framework \citep{bennington2022residual} and its extensions \citep{heimdal2026update, Heimdal2025}, alongside other methods which employ self-organizing maps \citep{landschutzer2016som}, ensemble feed-forward networks \citep{chau2022lsce}, extremely randomized trees \citep{wanninkhof2025ert}, and high-resolution target-transformation approaches \citep{gregor2024highres}.
Importantly, most existing approaches represent the predicted ocean state by a single point estimate. Existing uncertainty estimates are generally obtained indirectly by repeatedly perturbing model inputs or by comparing predictions across multiple independently developed reconstruction methods evaluated within model-ensemble test beds \citep{gloege2021testbed}. These procedures are valuable, but they do not directly learn the conditional distribution of $p\mathrm{CO}_2$ given the observed environmental variables. A generative model that learns a calibrated and samplable distribution, $p(y\mid x)$, would instead make uncertainty an intrinsic component of each prediction, helping to identify poorly constrained conditions and providing a building block for probabilistic inference in the reconstruction problem.

We investigate whether such a model can be obtained without abandoning the tabular, tree-based framework already widely used by the ocean carbon community. We use Treeffuser \citep{beltranvelez2024treeffuser}, which represents the score function of a conditional diffusion process using gradient-boosted trees. The model therefore retains the inductive biases that make tree ensembles effective for heterogeneous environmental data while learning a full, samplable conditional distribution $p(y\mid x)$ for each observation, represented directly in the native target space rather than in a learned latent embedding. Using the predictors structure and experimental setup of a state-of-the-art framework \citep{heimdal2026update}, we model $\Delta p\mathrm{CO}_2$ observations from $2012$ to $2022$ as a function of $13$ variables, evaluating both point-prediction accuracy and distributional calibration. Our \textbf{main contributions} are:

\begin{itemize}\itemsep2pt \parskip0pt \topsep2pt
    \item To our knowledge, the first application of tree-based conditional diffusion to learn samplable predictive distributions of $\Delta p\mathrm{CO}_2$, complementing ensemble- and error-propagation-based uncertainty estimates in existing products.
    \item Observation-specific uncertainty estimates evaluated using distributional calibration and prediction-interval coverage.
    \item A comparison with point (including a neural-network) and probabilistic tree-based baselines, showing comparable point-prediction accuracy, the lowest CRPS, and coverage closest to the nominal $68\%$ and $95\%$ levels.
\end{itemize}
\section{Data and task}
\label{sec:data}
 
\paragraph{Observations and Predictors.}
We use gridded SOCATv$2025$ \citep{bakker2025surface} $\pco$ monthly observations at $1^\circ\!\times\!1^\circ$ spatial resolution over 2012--2022. The data is tabular and sparse in space and time, with one record per grid cell and month. Following the state-of-the-art $p\mathrm{CO}_2$-Residual framework \citep{bennington2022residual}, each record contains 13 predictors (see Table~1 in \citep{bennington2022residual}): temperature, salinity and base-10 log-transformed chlorophyll concentration at sea-surface together with their corresponding anomalies, log-transformed mixed-layer depth, the atmospheric CO$_2$ trend, three unit-sphere spatial coordinates $(A,B,C)=(\sin\lambda,\ \sin\mu\cos\lambda,\ - \cos\mu\cos\lambda)$ derived from latitude $\lambda$ and longitude $\mu$, and two annual harmonics $(T_0,T_1)=\big(\cos\tfrac{2\pi d}{365},\ \sin\tfrac{2\pi d}{365}\big)$ of the day-of-year $d$ of the (mid-month) observation.

\paragraph{The $\Delta p\mathrm{CO}_2$ Target.}
We define as target variable $\Delta p\mathrm{CO}_2$, {\it i.e.} the difference between the fugacity of $\mathrm{CO}_2$ and the atmospheric $\mathrm{CO}_2$ mole fraction, following the approach of \citep{gregor2024highres, Fay2024, wanninkhof2025ert, heimdal2026update}:
\begin{equation*}
    \Delta p\mathrm{CO}_2 \, = \, \mathrm{fCO}_2 \, - \, \mathrm{xCO}_2.
\end{equation*}
In this definition, $\mathrm{xCO}_2$ varies with time, but has been averaged across the spatial dimensions. The use of $\Delta p\mathrm{CO}_2$ as a target aims to remove the long-term trend of ocean $\mathrm{fCO}_2$, alleviating the issue of modeling an evolving target.

\paragraph{Split.}
We use a temporally interleaved hold-out: within the global monthly sequence every fifth month is assigned to the test set and the remainder to training, so that test months are spread across all seasons and years (e.g.\ test months $\{1,6,11\}$ in 2012, $\{4,9\}$ in 2013, and so on). This evaluates temporal \emph{interpolation} between observed months rather than forward extrapolation, and yields ${28{,}879}$ test records. All models see identical train/test splits and identical predictors.
\begin{figure}[ht]
  \centering
  \includegraphics[width=\linewidth]{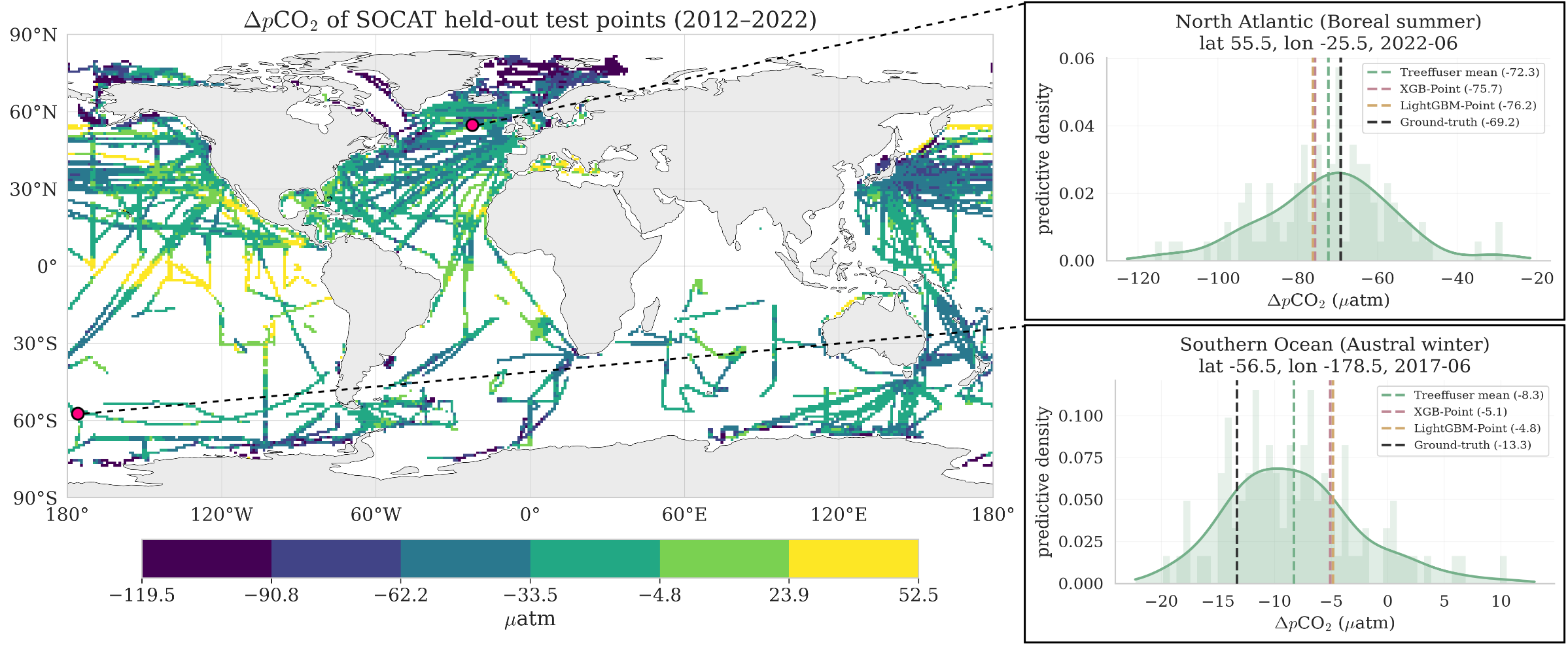}
  \caption{Point estimates versus the full predictive distribution of $\Delta p\mathrm{CO}_2$ at test-set locations. (Left) Spatial distribution of the held-out SOCAT observations used for evaluation. (Right) Treeffuser predictive distributions at the highlighted locations. Vertical lines indicate the Treeffuser predictive mean, the XGB-Point and LightGBM-Point predictions, and the observed SOCAT value.}
  %, shown as normalized histograms with overlaid kernel density estimates.
  \label{fig:headline}
\end{figure}

\section{Methods and metrics}
\label{sec:methods}

\paragraph{Benchmarked models.}
We focus on tree-based methods and group them by how they represent predictive uncertainty (full descriptions in Appendix~\ref{app:methods}). XGB-Point \citep{chen2016xgboost} and LightGBM-Point \cite{ke2017lightgbm} estimate only the conditional mean and yield no distribution; as a non-tree reference we also include ANN-Point, a feed-forward network matching the regression stage of SOM-FFN \citep{landschutzer2013neural}. XGB-Quantile fits a pinball-loss model on a grid of levels \citep{koenker1978regression}; XGB-Residual adds a single, input-independent Gaussian noise model to a point prediction; XGB-Ensemble uses the spread of bootstrap-fitted models, analogous to deep ensembles \citep{lakshminarayanan2017ensembles}; and NGBoost \citep{duan2020ngboost} predicts the parameters of a Gaussian via natural-gradient boosting. Treeffuser \citep{beltranvelez2024treeffuser} learns the conditional distribution of $\Delta p\mathrm{CO}_2$, using LightGBM as the score model. Samples are generated by integrating the reverse-time SDE, giving a samplable distribution with no fixed parametric family. Because its score model is LightGBM, we add a LightGBM-Point baseline to separate the effect of the tree library from that of the diffusion layer.

\paragraph{Fair-comparison protocol.}
All methods use the same training-test split, predictors, preprocessing, and evaluation observations. Where architectures permit, the tree-based methods also share hyperparameters and a fixed random seed; however, some differences remain as reported in Appendix~\ref{app:hparams}. In XGB-Ensemble, each member is fitted to a distinct bootstrap resample of the training data (drawn with replacement) using a different seed. For Treeffuser, we use $100$ draws from the predictive distribution and $50$ integration steps of the underlying SDE, selected by the analysis in Appendix~\ref{app:ablation}.

\paragraph{Metrics.}
We evaluate the predicted distributions using the continuous ranked probability score (CRPS), a strictly proper scoring rule for the full predictive distribution; empirical coverages (Cov68, Cov95) and the corresponding mean widths of the prediction intervals (Wid68, Wid95); and the $68\%$ and $95\%$ interval (Winkler) scores (IntScore68, IntScore95), which penalize both wide intervals and observations falling outside them \citep{gneiting2007scoring}. %The $68\%$ and $95\%$ prediction intervals correspond to the nominal 1$\sigma$ and 2$\sigma$ coverage levels, respectively, for a Gaussian distribution. 
Lower values indicate better performance for CRPS, the interval scores, and interval widths, whereas Cov68 and Cov95 are evaluated by their proximity to the corresponding nominal coverage levels of $68\%$ and $95\%$. Point predictions are evaluated using the Root Mean Squared Error (RMSE) and the coefficient of determination $R^2$. All metrics are expressed in \si{\micro atm}, except for Cov68, Cov95, and $R^2$, which are dimensionless.

\section{Results}
\label{sec:results}

\paragraph{Probabilistic accuracy and calibration.}
Table~\ref{tab:uq} summarizes probabilistic performance on the SOCAT held-out test set. Treeffuser achieves the lowest CRPS ($8.229$ \si{\micro atm}), the lowest $95\%$ interval score ($81.7$ \si{\micro atm}), and coverage closest to the nominal $68\%$ and $95\%$ levels ($0.686$ and $0.930$). XGB-Quantile is the closest competitor in CRPS ($8.339$ \si{\micro atm}) and has the lowest $68\%$ interval score ($45.6$ \si{\micro atm}), but substantially under-covers the observations ($0.560$ and $0.880$), indicating overconfident predictions. NGBoost provides the next-best coverage ($0.659$ and $0.927$) at the cost of wider intervals and higher CRPS and interval scores. XGB-Ensemble is the most overconfident, with narrow intervals covering only $0.230$ and $0.425$ and yielding the highest interval scores. Overall, Treeffuser offers the best balance of distributional accuracy and calibrated coverage.
 
\paragraph{The cost in point accuracy is modest.}
A natural concern is whether modeling the full predictive distribution comes at the expense of point-prediction accuracy. Table~\ref{tab:uq} shows that Treeffuser's predictive mean achieves an RMSE of $18.41$ \si{\micro atm} and $R^2=0.77$, compared with $17.73$ \si{\micro atm} and $R^2=0.79$ for XGB-Point. Since Treeffuser uses LightGBM as its base learner, we additionally include a LightGBM-Point model with matched hyperparameters. Its performance (RMSE $=17.82$ \si{\micro atm}, $R^2=0.78$) is close to that of XGB-Point, indicating that the modest reduction in point-prediction accuracy may not be primarily due to the type of base learner, but may instead reflect the additional objective of modeling the full predictive distribution. The neural-network baseline, ANN-Point, is less accurate than all tree-based point models (RMSE $=20.02$ \si{\micro atm}, $R^2=0.73$), supporting gradient-boosted trees as the backbone for this task. Overall, Treeffuser provides a calibrated, samplable predictive distribution while retaining competitive first-moment accuracy.

\paragraph{Per-observation uncertainty.} Figure~\ref{fig:headline} illustrates Treeffuser predictive distributions at two selected test grid points; at these locations the predictive mean lies closer to the observed $\Delta p\mathrm{CO}_2$ than the baseline point estimates, most visibly in the Southern Ocean, although this is not the case at every location. Each predictive distribution is conditional on the predictors of a single location-month and does not encode correlations between locations or months; the resulting per-observation uncertainty can be mapped and summarized by basin (Figures~\ref{fig:fig_a3}--\ref{fig:fig_a5}). Its monthly climatology reveals a physically interpretable seasonal structure, with lowest confidence in spring--summer of each hemisphere (Figures~\ref{fig:fig_a1} and~\ref{fig:fig_a2}).

\begin{table}[ht]
  \centering
  %\caption
  \captionof{table}{Probabilistic and point scores of candidate models on the SOCAT test set (N=28,879). For point models probabilistic scores are not defined. Best values in bold; for coverage, closest to nominal. Values in \si{\micro atm} except coverage and $R^2$.}

  \label{tab:uq}
  \small
  \setlength{\tabcolsep}{5pt}
    \begin{tabular}{lc|ccc|ccc|cc}
    
    \toprule
    Model & CRPS & Cov68 & Wid68 & IntScore68 & Cov95 & Wid95 & IntScore95 & RMSE & $R^2$ \\
    
    \midrule
    
    Treeffuser & \best{8.229} & \best{0.686} & 27.452 & 45.8 & \best{0.930} & 55.04 & \best{81.7} & 18.41 & 0.77 \\
    
    NGBoost & 9.717 & 0.659 & 31.015 & 53.0 & 0.927 & 61.13 & 91.2 & 20.34 & 0.72 \\
    
    XGB-Quantile & 8.339 & 0.560 & 20.771 & \best{45.6} & 0.880 & 50.38 & 93.0 & 17.74 & 0.79 \\
    
    XGB-Residual & 8.708 & 0.729 & 26.186 & 50.5 & 0.900 & 51.61 & 121.7 & 17.73 & 0.79 \\
    
    XGB-Ensemble & 9.190 & 0.230 & 7.575 & 55.1 & 0.425 & 14.85 & 225.1 & \textbf{17.37} & \textbf{0.80} \\

    \midrule
    XGB-Point & -- & -- & -- & -- & -- & -- & -- & 17.73 & 0.79  \\
    LightGBM-Point & -- & -- & -- & -- & -- & -- & -- & 17.82 & 0.78 \\
    ANN-Point & -- & -- & -- & -- & -- & -- & -- & 20.02 & 0.73 \\
    
    \bottomrule

    \end{tabular}
    
    %\\[3pt]
    %{\footnotesize Cov=coverage, Wid=mean width ($\mu$atm), IntScore=Winkler score.}

\end{table}

\section{Discussion}
\label{sec:discussion}
Observation-based $\Delta p\mathrm{CO}_2$ datasets are typically provided as gridded products, with uncertainty either omitted or represented through parametric or heuristic assumptions. Our results show that Treeffuser recovers a \emph{calibrated}, samplable predictive distribution of $\Delta p\mathrm{CO}_2$ for each observation while retaining almost all of the accuracy of XGB-Point. It attains the lowest CRPS and $95\%$ interval score in Table~\ref{tab:uq}, and the coverage closest to nominal $68\%$ and $95\%$. Calibrated observation-level uncertainty is a prerequisite for carbon-budget applications, where overconfidence would understate flux uncertainty, but it is not sufficient: propagating it to basin or global fluxes also requires the covariance between grid cells and months, which our model does not learn. The cost of obtaining a trustworthy distribution is small, an RMSE increase of $0.68\,\si{\micro atm}$ ($\approx 3.8\%$) and a drop of $0.02$ in $R^2$ relative to XGB-Point, which we regard as a favorable trade for a full predictive distribution, and may potentially be further improved by using a different score model for Treeffuser. 
 
\paragraph{Limitations.}
Treeffuser's benefits come at a cost. Sampling requires solving a reverse SDE, so inference is slower than a single tree pass ($897$\,s for the $28{,}879$ test records with $100$ draws and $50$ integration steps; Appendix~\ref{app:ablation}). Our evaluation also covers the easier of two regimes. Because test months are interleaved with training months, test records come from the same years, seasons, and SOCAT-sampled regions as the training data, so the model only interpolates in time between observed months. It does not test extrapolation, that is, prediction at locations SOCAT never sampled or in periods after the training data, where the predictors are supplied by Earth system or global ocean biogeochemistry models; this reconstruction regime is harder and more relevant operationally. Finally, each prediction is a conditional distribution $p(y \mid x)$ for a single location-month, so the model encodes no spatial or temporal covariance between predictions. Capturing it would require map-based approaches that depart from the tabular SOCAT setting.
 
\paragraph{Conclusion.}
We applied Treeffuser, a tree-based diffusion model, to learn the conditional distribution of $\Delta p\mathrm{CO}_2$ given $13$ environmental, spatial, and temporal predictors. A single trained model, evaluated at the predictors of each location-month, yields an observation-specific predictive distribution from which samples can be drawn. Rather than embedding observations into a latent space, this approach preserves the tabular structure of SOCAT and the tree-based paradigm already adopted by the ocean carbon community, replacing a single predicted value with a full conditional distribution. Among the methods tested, it produced the best-calibrated predictions at only a modest cost in first-moment accuracy. These results point to a practical way of equipping data-driven ocean carbon products with samplable, observation-level predictive uncertainty. Because each distribution describes one location-month independently of the others, combining the model with a representation of spatio-temporal dependence, and evaluating it on spatially held-out data, are the next steps toward probabilistic reconstruction of the sparse and biased observational field.

\begin{ack}
We acknowledge funding from NSF through the Learning the Earth with Artificial intelligence and Physics (LEAP) Science and Technology Center (STC) (Award \#2019625). VA acknowledges support from a PIVOT Research award (Award \#12871) from the Simons Foundation. GAM, THH and ARF acknowledge support from US funding agencies (NOAA NA24OARX431G0151-T1-01 and NSF OCE 2333608), from Schmidt Sciences, LLC for the COCO2 project, and the European Space Agency (ESA) through the Ocean Carbon for Climate (OC4C; 3-18399/24/I-NB) project. The Surface Ocean CO2 Observing Network (SOCONET), the Surface Ocean CO2 Atlas (SOCAT) and the Surface Ocean CO2 Mapping (SOCOM) program are linked international efforts that, along with additional critical data from satellites, models and re-analysis, enable the routine delivery of ocean carbon and carbonate system data products. The loss of any activity (SOCONET, SOCAT, or SOCOM) or data source (satellite, in situ, or re-analysis) would prevent delivery of data products to global or regional assessments (Global Carbon Budget, IPCC, and others) that guide policy decisions. SOCONET, SOCAT, and SOCOM are endorsed by the International Ocean Carbon Coordination Project (IOCCP, an Infrastructural Project of the Scientific Committee on Oceanic Research) and the Global Ocean Observing System (GOOS) Biogeochemistry Panel. The many researchers, funders, and agencies responsible for these activities and data are thanked for their contributions. We thank Maurie Keppens for her support in configuring the neural network used as one of the baseline models in this work.
\end{ack}

%\section*{References}
\clearpage
\bibliographystyle{unsrtnat}
\bibliography{references}

%%%%%%%%%%%%%%%%%%%%%%%%%%%%%%%%%%%%%%%%%%%%%%%%%%%%%%%%%%%%

\clearpage
\appendix
% Prefix a "A" to all equations, figures, tables and reset the counter 
\setcounter{equation}{0}
\setcounter{figure}{0}
\setcounter{table}{0}
% \setcounter{page}{1}
% \makeatletter
\renewcommand{\thetable}{\thesection\arabic{table}}
\renewcommand{\thefigure}{\thesection\arabic{figure}}
\renewcommand{\theequation}{\thesection\arabic{equation}}

\section{Additional Results}
\label{app:additional_results}
\setcounter{table}{0}
\setcounter{figure}{0}

\begin{table}[ht]
\centering
\caption{Treeffuser predictive standard deviation ($\sigma$) of the $\Delta p\mathrm{CO}_2$, by ocean basin (2012--2022). Most/least confident months are the climatological maxima/minima of the mean per-cell $\sigma$; MAD columns give the basin-monthly mean absolute deviation of the Treeffuser predictive mean from the XGB-Point prediction and from observations. Values in \si{\micro atm}.}
\label{tab:basin_sigma_full}
\begin{tabular}{lrrcccccc}
\toprule
 & & & \multicolumn{2}{c}{Most confident} & \multicolumn{2}{c}{Least confident} & \multicolumn{2}{c}{MAD vs.} \\
\cmidrule(lr){4-5}\cmidrule(lr){6-7}\cmidrule(lr){8-9}
Basin & $N$ & $\overline{\sigma}$ & Month & $\sigma_{\min}$ & Month & $\sigma_{\max}$ & XGB-Point & Obs. \\
\midrule
Atlantic & 10,497 & 16.89 & Jan & 11.98 & May & 22.68 & 0.87 & 1.31 \\
Pacific  & 13,588 & 15.13 & Feb & 10.91 & Jul & 18.65 & 0.68 & 1.78 \\
Southern &  3,561 & 16.80 & Sep & 10.05 & Jan & 19.15 & 1.27 & 3.52 \\
\bottomrule
\end{tabular}
\end{table}

\begin{figure}[ht]
  \centering
  \includegraphics[width=\linewidth]{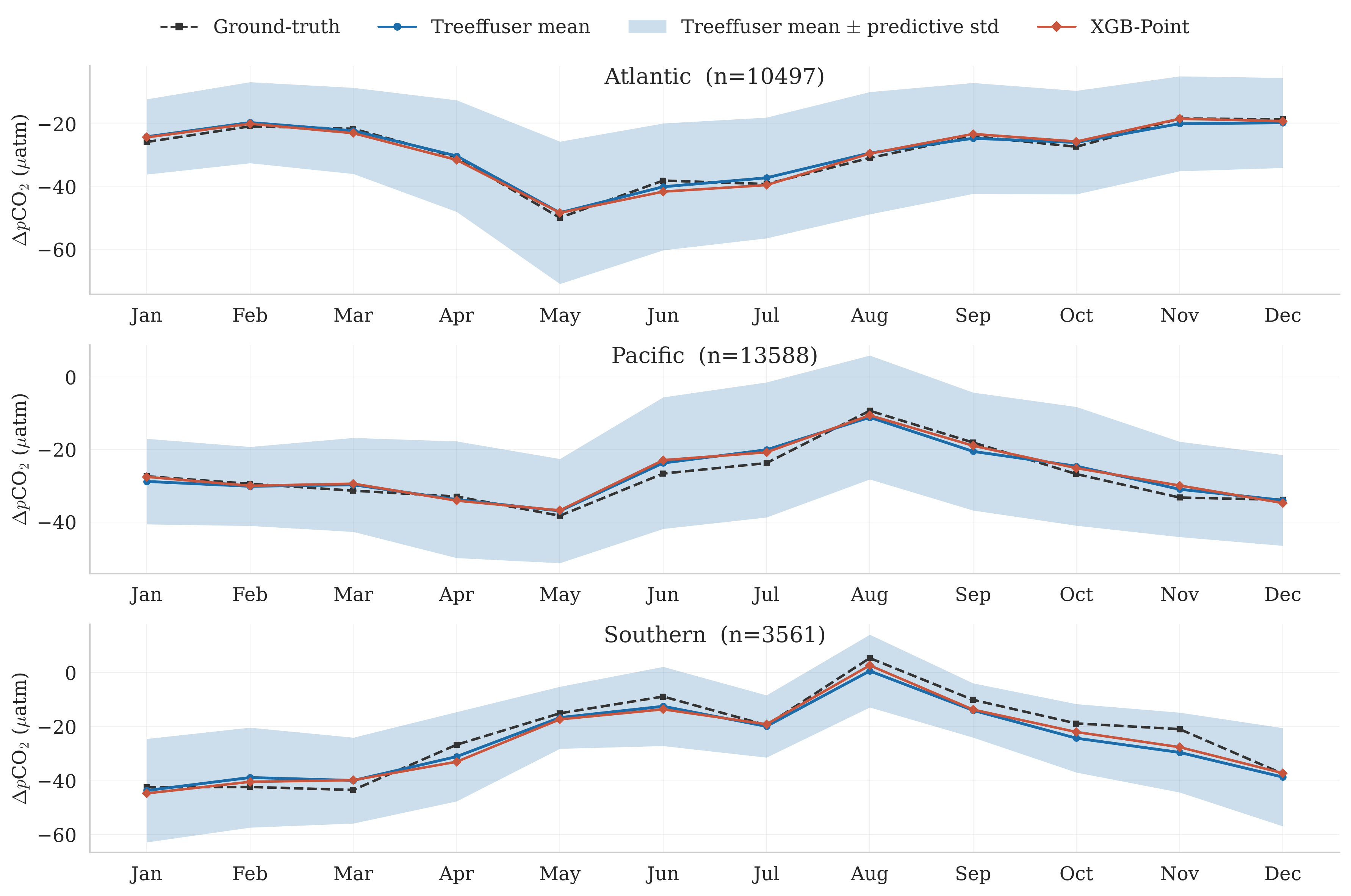}
  \caption{Monthly climatology of $\Delta p\mathrm{CO}_2$ across ocean basins (2012–2022). Solid line: Treeffuser predictive-mean, basin-averaged; shaded band: $\pm$ mean per-cell predictive standard deviation; dashed line: gridded observations; diamonds: XGB-Point. Axes are free per basin.}
  \label{fig:fig_a1}
\end{figure}

\begin{figure}[ht]
  \centering
  \includegraphics[width=\linewidth]{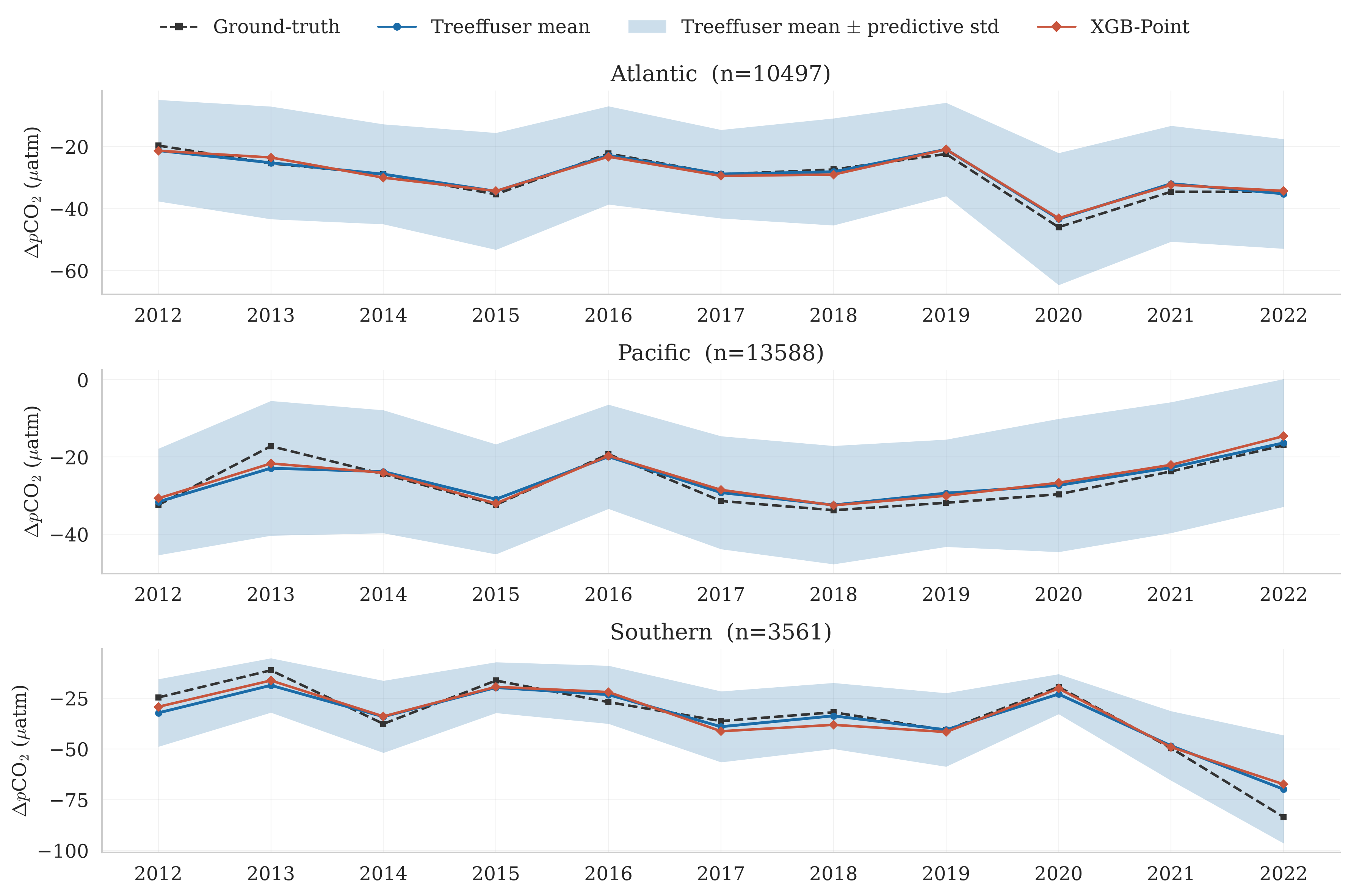}
  \caption{Interannual variation of $\Delta p\mathrm{CO}_2$ across ocean basins (2012--2022). Each point is an annual mean over all cells in the basin. Solid line: Treeffuser predictive-mean, basin-averaged; shaded band: $\pm$ mean per-cell predictive standard deviation; dashed line: gridded observations; diamonds: XGB-Point. Axes are free per basin.}
  \label{fig:fig_a2}
\end{figure}

\paragraph{Seasonal structure of predictive uncertainty.}
Figure~\ref{fig:fig_a1} shows the monthly climatology of $\Delta p\mathrm{CO}_2$ and its predictive uncertainty across ocean basins. Treeffuser closely follows both XGB-Point and the gridded observations: its basin-monthly mean absolute difference is at most \SI{1.27}{\micro atm} from XGB-Point and \SI{3.52}{\micro atm} from the observations (Table~\ref{tab:basin_sigma_full}), indicating that distributional modeling retains the central accuracy of the point model. Predictive uncertainty is nevertheless strongly seasonal. Basin-mean $\sigma$ values are similar (\SIrange{15.13}{16.89}{\micro atm}), but monthly values range from \SIrange{10.05}{11.98}{\micro atm} in the most-confident months to \SIrange{18.65}{22.68}{\micro atm} in the least-confident months. The least-confident months generally occur during spring or summer in the corresponding hemisphere, plausibly reflecting greater variability in the $\Delta p\mathrm{CO}_2$ associated with biological activity, stratification, and episodic mixing. Interannual variation is shown in Figure~\ref{fig:fig_a2}.

\begin{figure}[ht]
  \centering
  \includegraphics[width=\linewidth]{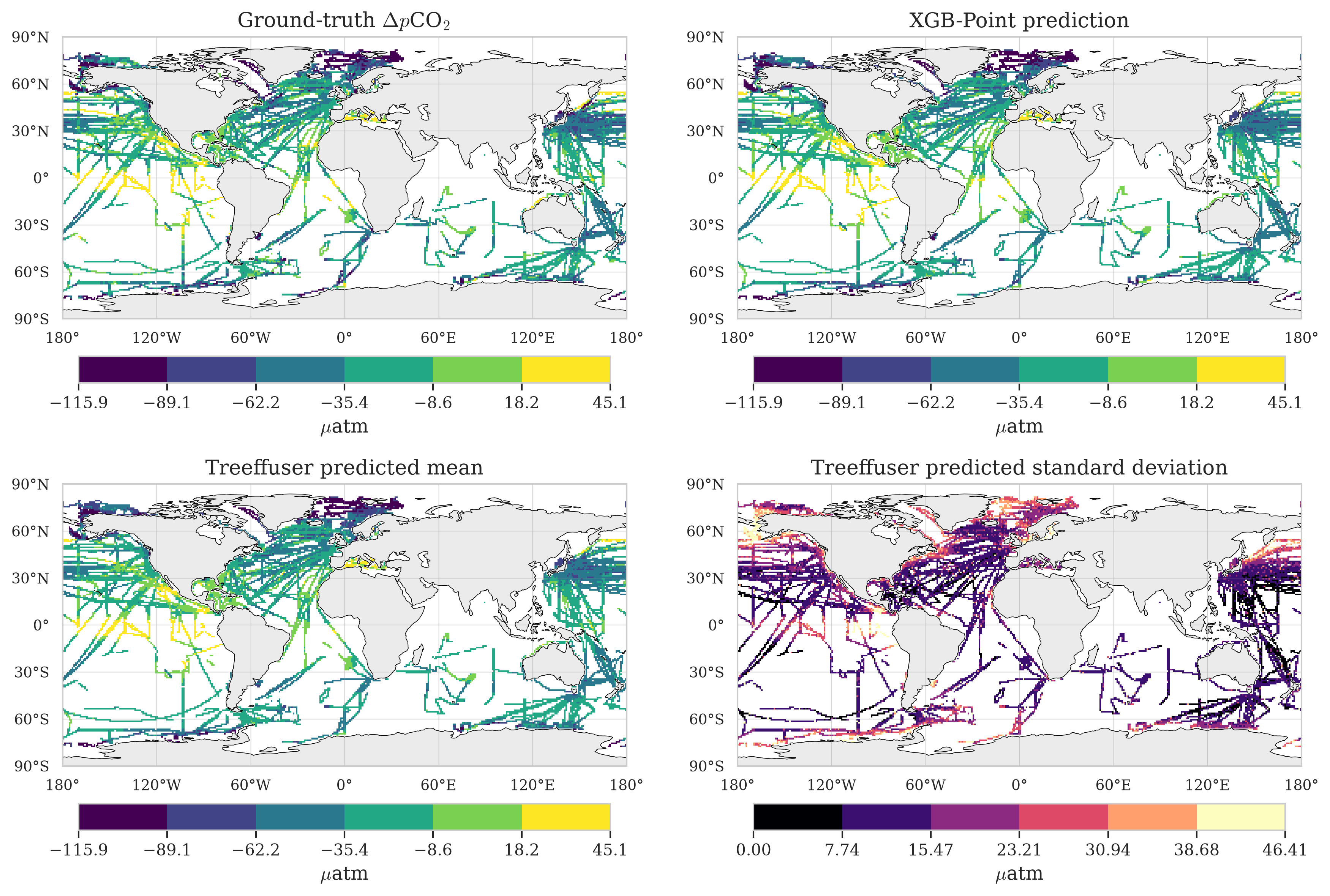}
  \caption{Global maps of test-set $\Delta p\mathrm{CO}_2$, temporally averaged per $1^\circ$ cell over 2012--2022. Top row: gridded observations (left) and the deterministic XGB-Point prediction (right). Bottom row: Treeffuser predictive mean (left) and per-cell predictive standard deviation (right). The three mean fields share a common colour scale, whereas the standard-deviation panel uses a separate scale (magma). Values in \si{\micro atm}.}
  \label{fig:fig_a3}
\end{figure}

\begin{figure}[ht]
  \centering
  \includegraphics[width=\linewidth]{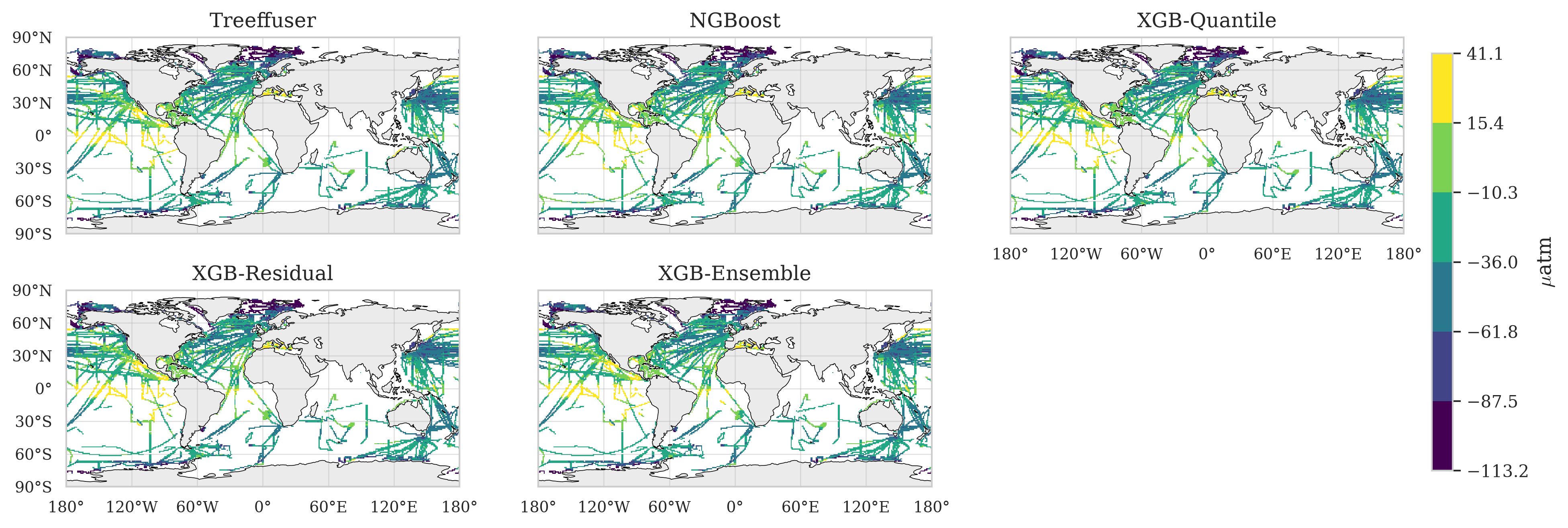}
  \caption{Predicted mean $\Delta p\mathrm{CO}_2$ by model, temporally averaged per grid point over 2012--2022.}
  \label{fig:fig_a4}
\end{figure}

\begin{figure}[ht]
  \centering
  \includegraphics[width=\linewidth]{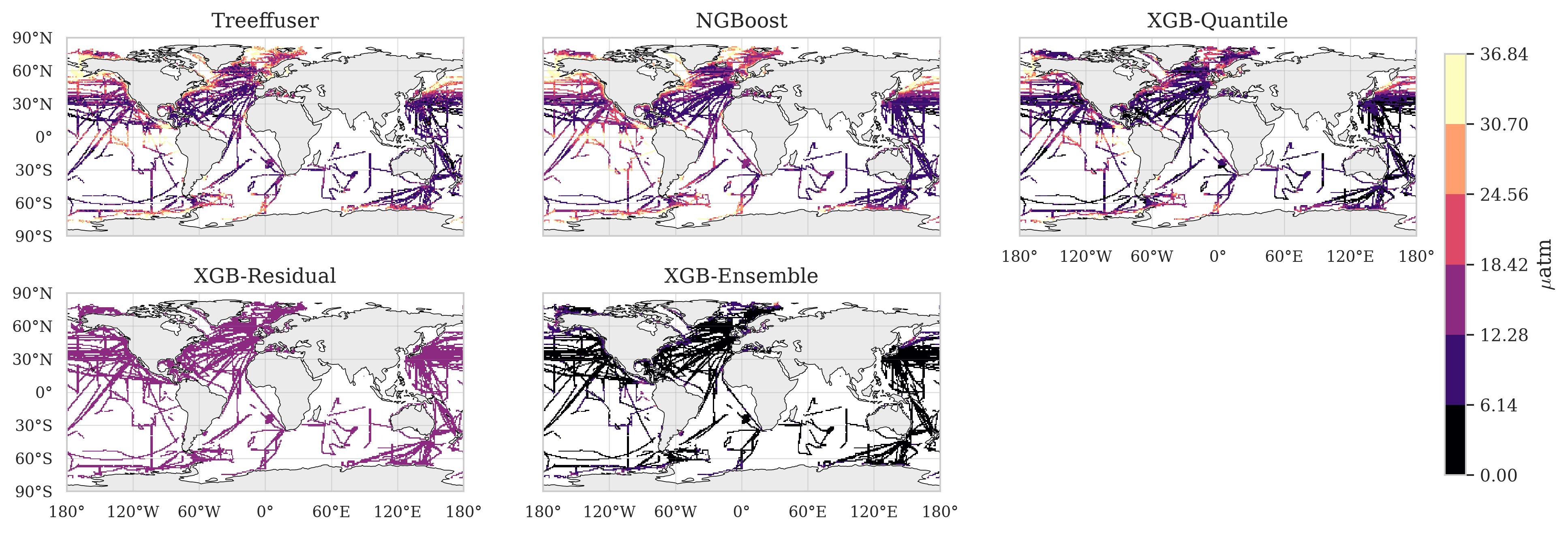}
  \caption{Predicted $\sigma$ of $\Delta p\mathrm{CO}_2$ by model, temporally averaged per grid point over 2012--2022.}
  \label{fig:fig_a5}
\end{figure}

\clearpage
\section{Model descriptions and training details}
\label{app:methods}
\setcounter{table}{0}

This appendix specifies every benchmarked method: its objective, how it is trained, the distributional object it produces, and the properties that explain its behaviour in Table~\ref{tab:uq}. We first fix notation and the shared protocol, then describe the deterministic baselines (XGB-Point, LightGBM-Point, and ANN-Point), the three XGB uncertainty variants, NGBoost, and finally the proposed method, Treeffuser.

\subsection{Notation and shared protocol}
\label{app:protocol}
Let $\mathcal{D}=\{(x_i,y_i)\}_{i=1}^{n}$ with covariates $x_i\in\mathbb{R}^{13}$ (the predictors of Section~\ref{sec:data}) and scalar target $y_i$ equal to the $\Delta \pco$. All methods estimate a functional of the conditional law $p(y\mid x)$. The deterministic baselines (XGB-Point, LightGBM-Point, and ANN-Point) estimate only the conditional mean $\mu(x)=\mathbb{E}[y\mid x]$. The remaining methods estimate quantiles, an ensemble spread, a parametric density, or the full density.

Every method uses the \emph{same} temporally interleaved train/test split (Section~\ref{sec:data}) and the \emph{same} $13$ predictors. The tree-based methods are trained on unscaled inputs, since tree ensembles are invariant to monotone feature transforms. The two exceptions are ANN-Point, which z-scores its inputs and target using training-set statistics because a neural network is sensitive to input scale (Appendix~\ref{app:ann}), and Treeffuser, which standardizes covariates and target internally before diffusion (Appendix~\ref{app:treeffuser}). All predictions are mapped back to \si{\micro atm} before scoring.

To isolate the effect of the uncertainty mechanism from that of tuning, all tree-based methods share the same boosting hyperparameters (\texttt{n\_estimators=1000}, \texttt{max\_depth=9}, \texttt{learning\_rate=0.05}) wherever the method admits them, and all use the same fixed random seed. These values come from a prior grid search of XGB-Point on normalized mean squared error and were not re-tuned for any individual method. The sole exception to the shared seed is XGB-Ensemble, whose members must differ and which therefore uses a grid of seeds (Appendix~\ref{app:ensemble}).

Metrics are reported in Table~\ref{tab:uq}. Point-accuracy metrics use each method's mean prediction (the median for XGB-Quantile). Probabilistic metrics are computed in closed form for the Gaussian methods (NGBoost, XGB-Residual), from the predicted quantile grid for XGB-Quantile, and from empirical samples for XGB-Ensemble and Treeffuser (Appendix~\ref{app:metrics}). Hyperparameters are collected in Tables~\ref{tab:hparam-treeffuser} and~\ref{tab:hparam-baselines}.

\subsection{XGB-Point: deterministic baseline}
\label{app:point}

XGB-Point is a gradient-boosted regression-tree model \citep{chen2016xgboost} trained by second-order (Newton) boosting under the squared-error loss. The predictor is an additive ensemble $\hat{\mu}(x)=\sum_{k=1}^{K} f_k(x)$ of CART trees $f_k$; at each round the new tree minimizes a second-order Taylor expansion of the loss augmented with a complexity penalty $\Omega(f)=\gamma L+\tfrac12\lambda\lVert w\rVert_2^2+\alpha\lVert w\rVert_1$ over the $L$ leaf weights $w$, and is added with shrinkage (learning rate) $\eta$. Row and column subsampling provide additional regularization. XGB-Point returns only the conditional mean and therefore carries no uncertainty; it plays two roles in this study: it is the point-prediction reference in Table~\ref{tab:uq}, and it is the shared tree backbone reused (with identical hyperparameters and seed) by XGB-Residual and XGB-Ensemble, and mirrored by XGB-Quantile.

\subsection{LightGBM-Point: deterministic baseline}
\label{app:lgbm}
LightGBM-Point is a gradient-boosted regression-tree model \citep{ke2017lightgbm} trained, like XGB-Point, by second-order boosting under the squared-error loss, yielding an additive ensemble $\hat{\mu}(x)=\sum_{k=1}^{K} f_k(x)$ of regression trees added with shrinkage $\eta$ and regularized by the same row- and column-subsampling controls. It differs from XGB-Point in the two implementation choices that characterize the library. First, continuous predictors are bucketed into a fixed number of histogram bins and candidate splits are searched over these bins rather than over sorted feature values, lowering the cost of split finding; the histogram learner further incorporates exclusive feature bundling and gradient-based one-side sampling, the accelerations introduced in \citep{ke2017lightgbm}. Second, trees are grown leaf-wise (best-first), at each step the leaf offering the largest loss reduction is split, rather than level-wise, here capped at the shared \texttt{max\_depth}. Under the shared hyperparameters (Section~\ref{app:protocol}) these differences leave the two libraries closely comparable in capacity. Like XGB-Point, LightGBM-Point returns only the conditional mean and therefore carries no uncertainty; it plays two roles in this study: it is a second deterministic point reference in Table~\ref{tab:uq}, and it exposes the gradient-boosting backbone on which Treeffuser learns its score function, so that the gap between its prediction and the Treeffuser predictive mean isolates the contribution of the probabilistic (diffusion) layer from the choice of tree library.

\subsection{ANN-Point: feed-forward neural network baseline}
\label{app:ann}
ANN-Point is a fully connected feed-forward neural network (FFN), the non-tree model family most widely used in observation-based $\pco$ reconstruction: it is the regression stage of the SOM-FFN method \citep{landschutzer2013neural,landschutzer2014recent} and of the UExP-FNN-U product \citep{ford2024comprehensive}. Those methods first cluster the ocean into biogeochemical provinces with a self-organizing map (SOM) and then fit one FFN per province. We reproduce only the regression stage and train a single global network on the same $13$ predictors and the same train/test split as all other methods. This mirrors the single global models used for the tree-based methods, so that differences reflect the model family rather than the regionalization. The predictor is
\begin{equation}
  \hat{\mu}(x)=w_2^{\top}\tanh\!\big(W_1\tilde{x}+b_1\big)+b_2,
\end{equation}
with one hidden layer of $H=210$ units, a hyperbolic-tangent (tangent-sigmoid) activation and a linear output, as in \citet{landschutzer2013neural,landschutzer2014recent}. Unlike the tree ensembles, a neural network is sensitive to the scale of its inputs. We therefore z-score each predictor ($\tilde{x}$) and the target using training-set statistics, and invert the target transform at prediction time.

Training follows the cited works where an equivalent exists. The training rows are randomly split $70/30$ into fitting and validation subsets \citep{ford2024comprehensive}. The squared-error loss is minimized without weight decay, and training stops once the validation loss has not improved for $6$ consecutive epochs. Early stopping is therefore the only regularizer, as in the original SOM-FFN implementations. %% CHECK: confirm the 70/30 split and the 6-epoch patience against the three papers
The width $H$ was selected from $\{30,60,\dots,300\}$ by validation error. %% CHECK: this search is not in the repo (config.toml only states it). Keep this sentence only if you have the log; otherwise write "H=210 was fixed a priori".
The original works train with Levenberg--Marquardt, which is not available in scikit-learn. We instead use Adam \citep{kingma2015adam} (learning rate $10^{-3}$, mini-batches of $512$, at most $1000$ epochs) via scikit-learn's \texttt{MLPRegressor}, with the shared seed. Like XGB-Point and LightGBM-Point, ANN-Point returns only the conditional mean and carries no uncertainty. It is included as the neural-network point reference in Table~\ref{tab:uq}. Its higher error (RMSE $20.02$ vs.\ $17.73\uatm$ for XGB-Point) is consistent with the use of gradient-boosted trees as the backbone of the probabilistic methods benchmarked here.

\subsection{XGB-Quantile}
\label{app:quantile}

XGB-Quantile keeps the boosting machinery of XGB-Point but replaces the loss with the pinball (quantile) loss, so that gradient boosting performs conditional quantile regression \citep{koenker1978regression,meinshausen2006qrf}. For a quantile level $\tau\in(0,1)$ and prediction $q$,
\begin{equation}
  \ell_\tau(y,q)=\big(\tau-\mathbf{1}\{y<q\}\big)\,(y-q),
\end{equation}
whose minimizer is the conditional $\tau$-quantile $q_\tau(x)$. We fit a grid of $39$ levels $\tau\in\mathcal{Q}=\{0.025,0.05,\dots,0.975\}$ jointly with XGBoost's native multi-quantile objective (\texttt{reg:quantileerror}), using a single model. The predicted quantile function is made monotone by sorting (rearrangement) to remove any quantile crossing. Central $95\%$ intervals use the grid levels $(0.025, 0.975)$ directly; central $68\%$ intervals use $(0.16, 0.84)$, obtained by linear interpolation between adjacent grid levels. XGB-Quantile is distribution-free and heteroscedastic, but it represents the predictive law only as finely as the quantile grid, does not model the tails beyond the extreme levels, and estimates each level independently; on our task it is sharp but under-covered (Table~\ref{tab:uq}), i.e.\ mildly overconfident.

\subsection{XGB-Residual}
\label{app:residual}
%XGB-Residual augments the XGB-Point mean predictor $\hat{\mu}(x)$ with a single, input-independent noise model. The standard deviation $\hat\sigma$ of the model's residuals $e_i=y_i-\hat{\mu}(x_i)$ is estimated once on held-out data ($20\%$ of the training records), and the predictive law is the homoscedastic Gaussian $\widehat F(\cdot\mid x)=\mathcal N\big(\hat{\mu}(x),\hat\sigma^2\big)$; its CRPS, and intervals therefore use the Gaussian closed forms (Appendix~\ref{app:metrics}). Because $\hat\sigma$ does not depend on $x$, the predictive spread is \emph{constant} across the ocean: the method captures the average noise level but cannot widen in genuinely more uncertain regions, which yields reasonable central coverage yet a poor interval score where the true spread varies with $x$ (Table~\ref{tab:uq}). It shares the backbone hyperparameters and seed of XGB-Point.

XGB-Residual augments the XGB-Point mean predictor $\hat{\mu}(x)$ with a single, input-independent Gaussian noise model. We first fit XGB-Point on a random $80\%$ of the training records and estimate the residual standard deviation $\hat\sigma$ from its residuals $e_i=y_i-\hat{\mu}(x_i)$ on the remaining $20\%$, which gives $\hat\sigma=13.17\uatm$. The mean predictor $\hat\mu(x)$ is then refit on the full training set with the shared hyperparameters and seed. The predictive law is the homoscedastic Gaussian $\widehat F(\cdot\mid x)=\mathcal N\big(\hat{\mu}(x),\hat\sigma^2\big)$, so its CRPS and prediction intervals use the Gaussian closed forms (Appendix~\ref{app:metrics}). Because $\hat\sigma$ does not depend on $x$, the predictive spread is \emph{constant} across the ocean. The method captures the average noise level but cannot widen in more uncertain regions. This yields reasonable central coverage but a poor interval score wherever the true spread varies with $x$ (Table~\ref{tab:uq}). Its point predictions are identical to those of XGB-Point.

\subsection{XGB-Ensemble (bootstrap ensemble)}
\label{app:ensemble}

XGB-Ensemble estimates uncertainty from the disagreement of $B=100$ XGB models trained on independent bootstrap resamples of the training rows, the tree analogue of deep ensembles \citep{lakshminarayanan2017ensembles}. Each member $b$ is trained on rows resampled with replacement under a distinct random seed, the ``seed grid'' that permutes the training set, while all boosting hyperparameters are held fixed and equal to the shared backbone. The predictive distribution is the empirical distribution of the member predictions $\{\hat{\mu}_b(x)\}_{b=1}^{B}$, from which we take the mean, standard deviation, and quantiles. Crucially, this spread reflects only \emph{epistemic} variance induced by resampling the finite training set; it does \emph{not} include the \emph{aleatoric} observation and process noise present in $\pco$. The intervals are therefore systematically too narrow, which is exactly the strong overconfidence observed in Table~\ref{tab:uq} ($95\%$ coverage of only $0.425$).

\subsection{NGBoost}
\label{app:ngboost}

NGBoost performs probabilistic regression by using natural-gradient boosting to learn the parameters of a parametric predictive distribution \citep{duan2020ngboost}. We use a Gaussian predictive law, $y\mid x\sim\mathcal{N}\big(\mu(x),\sigma(x)^2\big)$, with parameters $\theta(x)=\big(\mu(x),\log\sigma(x)\big)$. Parameterizing the scale through $\log\sigma$ ensures that $\sigma>0$. At each boosting stage, NGBoost fits a separate set of shallow regression trees for each distributional parameter. These learners are fitted to the natural gradient, $\mathcal{I}(\theta)^{-1}\nabla_\theta S(\theta,y)$, where $\mathcal{I}(\theta)$ is the Fisher information matrix and $S$ is a strictly proper scoring rule. We use the default log score, $S=-\log\mathcal{N}(y;\mu,\sigma^2)$, corresponding to maximum likelihood. A line search determines the step size at each stage, and the resulting update is scaled by the learning rate. The use of the natural gradient makes the optimization invariant to the particular parameterization of the predictive distribution and helps stabilize the joint estimation of $\mu$ and $\sigma$.
The resulting predictive distribution is Gaussian, allowing quantiles, coverage, and CRPS to be computed analytically. Because $\sigma(x)$ is learned as a function of the input, NGBoost can represent heteroscedastic uncertainty. Its main limitation is the Gaussian assumption: it cannot capture skewness, heavy tails, or multimodality, which limits its performance relative to the nonparametric Treeffuser on CRPS and interval score.

\subsection{Treeffuser (proposed)}
\label{app:treeffuser}

Treeffuser \citep{beltranvelez2024treeffuser} is a conditional score-based diffusion model \citep{song2021sde} whose score function is estimated with gradient-boosted trees \citep{ke2017lightgbm} rather than a neural network. It learns the full, nonparametric conditional density $p(y\mid x)$ and produces predictions by drawing samples from the learned conditional distribution. All quantities we report, including mean, per-cell standard deviation, intervals, CRPS, are computed from these samples.

\paragraph{Forward diffusion.}
A variance-exploding SDE (VESDE) progressively corrupts the target on $t\in[0,1]$,
\begin{equation}
  \mathrm{d}Y = \sqrt{2\,\sigma(t)\,\sigma'(t)}\;\mathrm{d}W,
  \qquad
  \sigma(t)=\sigma_{\min}\Big(\tfrac{\sigma_{\max}}{\sigma_{\min}}\Big)^{t},
\end{equation}
with $\sigma_{\min}=0.01$ and $\sigma_{\max}=20$. Because the drift is zero and the diffusion is state-independent, the perturbation kernel is Gaussian in closed form,
\begin{equation}
  p_t(y_t\mid y_0)=\mathcal{N}\!\big(y_0,\;(\sigma(t)^2-\sigma(0)^2)\big),
  \label{eq:kernel}
\end{equation}
and as $t\!\to\!1$ the marginal approaches the prior $p_1\approx\mathcal{N}(0,\sigma_{\max}^2)$, from which sampling starts. The covariates $x$ are held fixed throughout and never diffused.

\paragraph{Score parameterization.}
Treeffuser models the conditional (Stein) score $s(y_t,x,t)\approx\nabla_{y}\log p_t(y_t\mid x)$ as
\begin{equation}
\label{eq:a4}
  s(y_t,x,t)=\frac{g(y_t,x,t)}{\sigma(t)},
\end{equation}
where $g$ is a single LightGBM regressor whose input is the concatenation $[\,y_t,\;x,\;t\,]$: the perturbed target, the covariates, and the diffusion time entered directly as an ordinary tabular feature. Feeding $t$ as a plain feature removes the need for the sinusoidal time embeddings used by neural score networks, and lets the model inherit the strengths of gradient-boosted trees, native handling of heterogeneous, skewed, and missing tabular inputs, scale invariance, and fast training.

\paragraph{Training objective (denoising score matching).}
The score is fit by denoising score matching (DSM) \citep{vincent2011connection}. Each training pair is replicated $R$ times ($R=30$); for every replicate we draw a time $t\sim\mathcal{U}(\varepsilon,1)$ ($\varepsilon=10^{-5}$) and noise $z\sim\mathcal{N}(0,1)$, form the perturbed target $y_t=y_0+\sigma(t)\,z$ using~\eqref{eq:kernel}, and regress $g$ onto the target $-z$ under squared loss:
\begin{equation}
  \min_{g}\;\mathbb{E}_{t,\,y_0,\,z}\,
  \big\lVert\, g(y_t,x,t)-(-z)\,\big\rVert_2^2 .
\end{equation}
This is the standard weighted DSM objective, with weighting $\sigma(t)^2$. Its minimizer satisfies $s=g/\sigma(t)\approx\nabla_y\log p_t(y_t\mid x)$. We fit one LightGBM model per output dimension. The boosting configuration follows XGB-Point, with \texttt{n\_estimators=1000}, \texttt{max\_depth=9}, and \texttt{learning\_rate=0.05}. We additionally set \texttt{num\_leaves=127} and \texttt{early\_stopping\_rounds=10}; all other hyperparameters are left at their default values (Table~\ref{tab:hparam-treeffuser}).

\paragraph{Sampling.}
To draw from $p(y\mid x)$ we integrate the reverse-time SDE \citep{anderson1982reverse}, which for the VESDE (zero drift, diffusion $g(t)=\sqrt{2\sigma\sigma'}$) is
\begin{equation}
  \mathrm{d}Y = -\,2\,\sigma(t)\,\sigma'(t)\;s(Y,x,t)\;\mathrm{d}t
  \;+\;\sqrt{2\,\sigma(t)\,\sigma'(t)}\;\mathrm{d}\bar W,
\end{equation}
started at $Y_1\sim\mathcal{N}(0,\sigma_{\max}^2)$ and integrated from $t=1$ to $t\!\approx\!0$ with the Euler--Maruyama scheme in $S$ steps. Repeating the integration $M$ times per input yields $M$ conditionally i.i.d.\ samples, whose empirical mean (or median) is the point prediction, whose standard deviation is the per-cell uncertainty, and whose empirical quantiles give prediction intervals. We use $M=100$ draws and $S=50$ integration steps, selected by the ablation in Appendix~\ref{app:ablation}.

\paragraph{Characteristics.}
Treeffuser is the only method in the benchmark that is simultaneously \emph{nonparametric} (it assumes no functional form for $p(y\mid x)$ and can represent skew, heavy tails, heteroscedasticity, and multimodality), \emph{samplable}, and \emph{tree-based}. The price is inference cost: producing $M$ samples requires $M \times S$ score (LightGBM) evaluations per input, so prediction is substantially slower than the single tree pass of the baselines (Appendix~\ref{app:ablation} quantifies the accuracy/compute trade-off).

\paragraph{Representation without a latent space.} Unlike neural score models, Treeffuser never maps its inputs into a learned embedding. Two properties follow from its construction. First, the covariates $x$ enter the score model directly as tabular features (Eq.~\ref{eq:a4}): the diffusion time $t$ is appended as an ordinary column and the inputs are never transformed into intermediate activations, so there is no continuous latent vector whose geometry encodes the data. Second, the diffusion process operates directly in the native target space: noise is added to $\Delta p\mathrm{CO}_2$ itself (Eq.~\ref{eq:kernel}), rather than to an encoded representation, as in latent diffusion models \citep{rozet2026lost} or approaches that perform generation in alternative representation spaces \citep{accarino2026wavelet}. What Treeffuser learns is therefore the conditional density $p(y\mid x)$ itself, carried by the score $\nabla_{y}\log p_t(y_t\mid x)$ across noise scales, rather than a geometry of embeddings. The tree ensemble still induces a representation of the inputs, the axis-aligned partition of covariate space defined by its leaves, but this is a discrete partition of physically meaningful variables, not a smooth latent with learned distances, and it is not where the generative process takes place. This keeps the representation interpretable in the original coordinates (each split is a threshold on an observable such as sst or chlorophyll), at the cost of transferability: the learned density is specific to the $\Delta p\mathrm{CO}_2$ task and, unlike a reusable embedding, does not transfer to other targets. Each predictive distribution is conditional on the predictors of a single location-month; the model does not represent dependence between grid cells or months.

\clearpage
\subsection{Hyperparameters}
\label{app:hparams}

\begin{table}[ht]
  \centering
  \caption{Treeffuser configuration (LightGBM score model, VESDE, and sampler).
  Values fixed by the implementation used in this work.}
  \label{tab:hparam-treeffuser}
  \small
  \begin{tabular}{lll}
    \toprule
    Component & Hyperparameter & Value \\
    \midrule
    \multirow{13}{*}{LightGBM score model}
      & n\_repeats               & 30        \\
      & n\_estimators            & 1000      \\
      & early\_stopping\_rounds  & 10        \\
      & eval\_percent            & 0.1       \\
      & num\_leaves              & 127       \\
      & max\_depth               & 9         \\
      & learning\_rate           & 0.05      \\
      & max\_bin                 & 255       \\
      & subsample\_for\_bin      & 200000    \\
      & min\_child\_samples      & 20        \\
      & subsample                & 1         \\
      & subsample\_freq          & 0         \\
      & seed                     & 380       \\
    \midrule
    \multirow{2}{*}{VESDE}
      & $\sigma_{\min}$, $\sigma_{\max}$ & 0.01, 20 \\
      & horizon $T$                & 1 \\
    \midrule
    \multirow{3}{*}{Training / sampling}
      & DSM repeats $R$            & 30 \\
      & number of integration steps $S$      & (see Appendix~\ref{app:ablation})  \\
      & number of draws $M$                  & (see Appendix~\ref{app:ablation}) \\
    \bottomrule
  \end{tabular}
\end{table}

\begin{table}[ht]
  \centering
  \caption{Hyperparameters for point-prediction models and models providing uncertainty estimates. These settings ensure a fair comparison, as described in Section~\ref{sec:methods}. Values not reported in the table are set to their default values.}
  \label{tab:hparam-baselines}
  \small
  \begin{tabular}{lll}
    \toprule
    Method & Hyperparameter & Value \\
    \midrule
    \multirow{4}{*}{\makecell[lt]{XGB backbone \\(Point/Residual/Ensemble/Quantile)}}
      & n\_estimators           & 1000 \\
      & max\_depth              & 9 \\
      & learning\_rate ($\eta$) & 0.05 \\
      & seed                    & 380 \\
    \midrule
    \multirow{4}{*}{LightGBM-Point}
      & n\_estimators           & 1000 \\
      & max\_depth              & 9 \\
      & learning\_rate ($\eta$) & 0.05 \\
      & num\_leaves             & 127 \\
    \midrule
    \multirow{9}{*}{ANN-Point}
      & hidden layers                 & 1 $\times$ 210 units \\
      & activation / output           & tanh / linear \\
      & optimizer                     & Adam \\
      & learning rate                 & $10^{-3}$ \\
      & batch size                    & 512 \\
      & L2 penalty ($\alpha$)         & 0 \\
      & early stopping                & 30\% validation, patience 6 \\
      & max epochs                    & 1000 \\
      & input / target scaling        & z-score (training statistics) \\
    \midrule
    XGB-Quantile & quantile grid $\mathcal{Q}$
      & \texttt{np.linspace(0.025, 0.975, 39)} \\
    \midrule
    \multirow{2}{*}{XGB-Ensemble} 
    & n\_members ($B$)               & 100 \\
    & per member seed           & \texttt{range(0, $B$)} \\
    \midrule
    \multirow{4}{*}{NGBoost}
      & n\_estimators      & 1000 \\
      & learning\_rate     & 0.05 \\
      & distribution & Normal   \\
      & score        & LogScore \\

    \bottomrule
  \end{tabular}
\end{table}

\clearpage
\section{Treeffuser configuration: draws $\times$ steps ablation}
\label{app:ablation}
\setcounter{table}{0}
We select $100$ draws and $50$ integration steps for Treeffuser. CRPS (fair estimator, comparable across draw counts) remains essentially flat across configurations ($8.214$--$8.245$). Empirical coverage approaches the nominal levels as the number of draws increases, mainly because more draws resolve the tails of the empirical quantiles; doubling the integration steps to $100$ lowers CRPS slightly but does not improve coverage. This sensitivity analysis was run on the test set; since all configurations differ by at most $0.031$\,\si{\micro atm} in CRPS and $0.028$ in coverage, the choice does not affect our conclusions. The selected configuration ($100\times50$) takes $897$\,s.
 
\begin{table}[ht]
  \centering
  \caption{Treeffuser ablation over predictive draws and integration steps of the underlying SDE. Time is reported in seconds, and the selected configuration in \textbf{bold}.}
  \label{tab:ablation}
  \small
  \begin{tabular}{ccrrrrrr}
  \toprule
    Draws & Steps & CRPS & Cov68 & Wid68 & Cov95 & Wid95 & Time (s) \\
    \midrule
    50 & 50 & 8.245 & 0.674 & 27.045 & 0.914 & 52.964 & 447 \\
    \best{100} & \best{50} & \best{8.229} & \best{0.686} & \best{27.452} & \best{0.930} & \best{55.042} & \best{897} \\
    200 & 50 & 8.236 & 0.692 & 27.634 & 0.936 & 55.945 & 1804 \\
    300 & 50 & 8.234 & 0.693 & 27.717 & 0.939 & 56.367 & 2689 \\
    \midrule
    50 & 100 & 8.221 & 0.670 & 26.796 & 0.911 & 52.405 & 1005 \\
    100 & 100 & 8.214 & 0.681 & 27.256 & 0.927 & 54.649 & 1990 \\
    200 & 100 & 8.221 & 0.688 & 27.450 & 0.934 & 55.579 & 3959 \\
    300 & 100 & 8.221 & 0.690 & 27.526 & 0.937 & 55.993 & 5992 \\
    \bottomrule
  \end{tabular}
  %\begin{tabular}{cccccc}
  %  \toprule
  %  Draws & Steps & CRPS & Cov95 & Wid95 & Time (s) \\
  %  \midrule
  %  50  & 50  & 8.245 & 0.914 & 52.96 & 447 \\
  %  \best{100} & \best{50}  & \best{8.229} & \best{0.930} & \best{55.04} & \best{897} \\
  %  200 & 50  & 8.236 & 0.936 & 55.95 & 1804 \\
  %  300 & 50  & 8.234 & 0.939 & 56.37 & 2689 \\
  %  \midrule
  %  50  & 100 & 8.221 & 0.911 & 52.41 & 1005 \\
  %  100 & 100 & 8.214 & 0.927 & 54.65 & 1990 \\
  %  200 & 100 & 8.221 & 0.934 & 55.58 & 3959 \\
  %  300 & 100 & 8.221 & 0.937 & 55.99 & 5992 \\
  %  \bottomrule
  %\end{tabular}
\end{table}

\section{Evaluation metrics and how they are computed per method}
\label{app:metrics}
\setcounter{table}{0}
We evaluate two aspects of each reconstruction: the accuracy of its point prediction and the quality of its predictive distribution. The subtlety is that the benchmarked methods do not expose the same distributional object, some return empirical samples, one returns a closed-form Gaussian, and one returns a quantile function, so each scalar metric must be computed differently for each method. This appendix gives, for every metric, its definition and the recipe used for each predictive object; Table~\ref{tab:metric-recipes} summarizes the recipes. CRPS is a strictly proper scoring rule and the interval score is a proper scoring rule \citep{gneiting2007scoring}; coverage, interval width, RMSE, and $R^2$ are complementary diagnostics. All metrics are averaged over the N=${28{,}879}$ test points and reported in \si{\micro atm}, except coverage and $R^2$, which are dimensionless.

%Unless noted, every score is a strictly proper scoring rule \citep{gneiting2007scoring} and is reported as an average over the $N=28,879$ test points, in \si{\micro atm} where the target has units.
 
\subsection{Notation and predictive objects}
\label{app:metrics-notation}
 
For test point $i$ let $y_i$ denote the observed $\Delta \pco$ and $F_i$ the method's predictive cumulative distribution function (CDF), with quantile function $F_i^{-1}$. Depending on the method, $F_i$ is available as one of:
\begin{itemize}\itemsep1pt \parskip0pt \topsep2pt
  \item \textbf{Samples} $\{y_i^{(s)}\}_{s=1}^{M}$ drawn from $F_i$: Treeffuser ($M=100$ reverse-SDE draws) and XGB-Ensemble ($M=B=100$ bootstrap members). The empirical CDF is $\widehat F_i(z)=\tfrac1M\sum_{s}\mathbf{1}\{y_i^{(s)}\le z\}$.
  \item \textbf{A parametric Gaussian} $F_i=\mathcal{N}(\mu_i,\sigma_i^2)$: NGBoost, with input-dependent $\sigma_i=\sigma(x_i)$, and XGB-Residual, with $\mu_i=\hat\mu(x_i)$ and constant $\sigma_i\equiv\hat\sigma$. Here $\Phi,\phi$ denote the standard normal CDF and PDF.
  \item \textbf{A quantile function on a grid} $\{q_{i,\tau}\}_{\tau\in\mathcal Q}$ with $q_{i,\tau}=F_i^{-1}(\tau)$: XGB-Quantile (sorted to enforce monotonicity; linearly interpolated between grid levels).
  %\item \textbf{A single interval} $[\ell_i,u_i]$ with a marginal coverage guarantee, but no full CDF: XGB-Conformal.
\end{itemize}
XGB-Point, LightGBM-Point, and ANN-Point return only a conditional mean and hence participate in the point-accuracy metrics alone (Table~\ref{tab:uq}). %Where a probabilistic object is absent (XGB-Point, and XGB-Conformal for distribution-level scores), the corresponding entries are undefined and marked with a dash in Table~\ref{tab:uq}.
 
\subsection{Point-accuracy metrics}
\label{app:metrics-point}
Let $\hat y_i$ be the method's point summary: the predictive mean for the sample-based and Gaussian methods (Treeffuser, XGB-Ensemble, NGBoost, XGB-Residual), and the predictive median $q_{i,0.5}$ for XGB-Quantile.

\begin{align}
  \mathrm{RMSE} &= \Big(\tfrac1n\textstyle\sum_{i}(y_i-\hat y_i)^2\Big)^{1/2}, &
  %\mathrm{MAE}  &= \tfrac1n\textstyle\sum_{i}\lvert y_i-\hat y_i\rvert, \\
  %\mathrm{Bias} &= \tfrac1n\textstyle\sum_{i}(\hat y_i-y_i), &
  R^2 &= 1-\frac{\sum_i (y_i-\hat y_i)^2}{\sum_i (y_i-\bar y)^2}.
\end{align}
RMSE is lower-is-better, whereas $R^2$, the fraction of variance explained, is higher-is-better. Scores are computed on $\Delta p\mathrm{CO}_2$ in \si{\micro atm}. 

%Scores are computed directly on the residual target the models predict (in \si{\micro atm}); we do not add back the thermal component. %This requires no correction for comparability: because $\pco^{T}$ is a deterministic additive term in the same units, it cancels in every prediction--observation difference, so RMSE, MAE, Bias---and, below, CRPS, interval width, and the interval score---are numerically identical whether evaluated on the residual or on the reconstructed $\pco=\hat y+\pco^{T}$. The one exception is $R^2$, whose denominator is the target variance: since $\mathrm{Var}(\pco)>\mathrm{Var}(\text{residual})$, the values reported here are the (more conservative) $R^2$ of the residual and would be larger if computed on full $\pco$.
 
\subsection{Continuous ranked probability score (CRPS)}
\label{app:metrics-crps}
 
CRPS is our headline probabilistic score. For a predictive CDF $F$ and observation $y$,
\begin{equation}
  \mathrm{CRPS}(F,y)=\int_{-\infty}^{\infty}\big(F(z)-\mathbf{1}\{z\ge y\}\big)^2\,\mathrm{d}z
  \;=\;\mathbb{E}_F\lvert X-y\rvert-\tfrac12\,\mathbb{E}_F\lvert X-X'\rvert,
  \label{eq:crps}
\end{equation}
where $X,X'\sim F$ are independent; the second (energy) form is the basis of the sample estimator. CRPS generalizes the mean absolute error to full distributions and is minimized only by the true conditional law, rewarding calibration and sharpness jointly. It is lower-is-better. We compute it as follows.
 
\paragraph{Sample-based methods (Treeffuser, XGB-Ensemble).}
Using the $M$ samples $\{y_i^{(s)}\}$,
\begin{equation}
  \widehat{\mathrm{CRPS}}(F_i,y_i)=
  \frac1M\sum_{s=1}^{M}\lvert y_i^{(s)}-y_i\rvert
  \;-\;\frac{1}{2M(M-1)}\sum_{s=1}^{M}\sum_{s'=1}^{M}\lvert y_i^{(s)}-y_i^{(s')}\rvert,
  \label{eq:crps-sample}
\end{equation}
which is unbiased for any $M$ \citep{jordan2019scoringrules}, so scores obtained with different numbers of draws are directly comparable.
 
\paragraph{Gaussian methods (NGBoost, XGB-Residual; closed form).}
With $\omega_i=(y_i-\mu_i)/\sigma_i$ \citep{gneiting2007scoring},
\begin{equation}
  \mathrm{CRPS}\big(\mathcal N(\mu_i,\sigma_i^2),y_i\big)=
  \sigma_i\Big[\,\omega_i\big(2\Phi(\omega_i)-1\big)+2\phi(\omega_i)-\tfrac{1}{\sqrt\pi}\,\Big].
\end{equation}
 
\paragraph{XGB-Quantile (quantile function).}
Using the quantile (pinball) representation of CRPS, $\mathrm{CRPS}(F,y)=2\int_0^1 \Lambda_\tau\!\big(F^{-1}(\tau),y\big)\,\mathrm{d}\tau$ with pinball loss $\Lambda_\tau(q,y)=\big(\mathbf{1}\{y\le q\}-\tau\big)(q-y)$, we integrate numerically over the fitted grid $\mathcal Q$,
\begin{equation}
  \widehat{\mathrm{CRPS}}(F_i,y_i)=\frac{2}{\lvert\mathcal Q\rvert}\sum_{\tau\in\mathcal Q}
  \Lambda_\tau\!\big(q_{i,\tau},y_i\big),
\end{equation}
i.e.\ twice the mean pinball loss over the $39$ equally weighted grid levels, an approximation of the integral over $\tau\in(0,1)$ that does not extend into the tails beyond $0.025$ and $0.975$.

%with trapezoidal weights $w_\tau$; the approximation improves as $\mathcal Q$ is refined.
 
\subsection{Prediction-interval coverage and width}
\label{app:metrics-cov}
 
For a central $(1-\alpha)$ interval we use the predictive quantiles $\ell_i=F_i^{-1}(\alpha/2)$ and $u_i=F_i^{-1}(1-\alpha/2)$, and report the prediction-interval coverage probability (PICP) and mean width (MPIW),
\begin{equation}
  \mathrm{Cov}_{1-\alpha}=\frac1n\sum_{i}\mathbf{1}\{\ell_i\le y_i\le u_i\},
  \qquad
  \mathrm{Wid}_{1-\alpha}=\frac1n\sum_{i}(u_i-\ell_i),
\end{equation}
at $\alpha\in\{0.32,0.05\}$ (the $68\%$ and $95\%$ levels). Coverage is best closest to its nominal level $1-\alpha$; width (sharpness) is lower-is-better \emph{given} adequate coverage; a narrow interval that under-covers is overconfident, not better. The endpoints $(\ell_i,u_i)$ are obtained per method as: the empirical $\alpha/2$ and $1-\alpha/2$ sample quantiles (sample-based methods); $\mu_i\pm z_{1-\alpha/2}\,\sigma_i$ with $z_{1-\alpha/2}=\Phi^{-1}(1-\alpha/2)$ (NGBoost and XGB-Residual); the predicted grid quantiles $q_{i,\alpha/2},q_{i,1-\alpha/2}$ (XGB-Quantile, linearly interpolated at $0.16$ and $0.84$ for the $68\%$ interval).
 
\subsection{Interval (Winkler) score}
\label{app:metrics-is}
 
To score an interval jointly for width and calibration we use the interval score \citep{winkler1972,gneiting2007scoring}. For the central $(1-\alpha)$ interval $[\ell_i,u_i]$,
\begin{equation}
  \mathrm{IS}_\alpha(\ell_i,u_i;y_i)=(u_i-\ell_i)
  +\frac{2}{\alpha}(\ell_i-y_i)\,\mathbf{1}\{y_i<\ell_i\}
  +\frac{2}{\alpha}(y_i-u_i)\,\mathbf{1}\{y_i>u_i\},
\end{equation}
reported at $\alpha=0.32$ (\textsf{IntScore68}) and $\alpha=0.05$ (\textsf{IntScore95}), averaged over the test set. It penalizes width plus a coverage penalty of $2/\alpha$ per unit of exceedance for observations that fall outside the interval, and is lower-is-better. It is defined for every method that produces an interval. The endpoints are those of Appendix~\ref{app:metrics-cov}.

\subsection{Summary of metric computation by method}
\label{app:metrics-summary}
 
\begin{table}[h]
  \centering
  \caption{How each metric is computed from each method's predictive object.
  ``samples'' denotes the empirical estimator over $M$ draws; ``Gaussian'' the
  closed form; ``grid'' the quantile/pinball route; ``--'' undefined.
  $\hat y$ is the point summary used for the accuracy metrics.}
  \label{tab:metric-recipes}
  \small
  \setlength{\tabcolsep}{4.5pt}
  \begin{tabular}{lccccc}
    \toprule
    Method & Predictive object & $\hat y$ & Interval $[\ell,u]$ & CRPS \\
    \midrule
    XGB-Point       & mean only            & mean            & --                 & --              \\
    LightGBM-Point  & mean only            & mean            & --                 & --              \\
    ANN-Point       & mean only            & mean            & --                 & --          \\
    XGB-Quantile    & quantile grid        & median          & grid quantiles     & grid      \\
    XGB-Residual    & Gaussian $(\hat\mu,\hat\sigma)$ & $\hat\mu$ & $\hat\mu\pm z\hat\sigma$ & Gaussian \\
    XGB-Ensemble    & members $\{\hat\mu_b\}$   & member mean & member quantiles   & samples   \\
    NGBoost         & Gaussian $(\mu,\sigma)$   & $\mu$      & $\mu\pm z\sigma$   & Gaussian  \\
    Treeffuser      & samples $\{y^{(s)}\}$ & sample mean    & sample quantiles   & samples \\
    \bottomrule
  \end{tabular}
\end{table}
 
%\subsection{Implementation details}
%\label{app:metrics-impl}
 
%Probabilistic scores for the sample-based methods use $M=100$ draws (matching the Treeffuser configuration of Appendix~\ref{app:ablation}); larger $M$ reduces the Monte-Carlo variance of~\eqref{eq:crps-sample} and of the empirical quantiles at $O(M^{-1/2})$. Interval endpoints and PIT for XGB-Quantile use the fitted grid $\mathcal Q$ \todo{state the grid}; the MACE grid is $\mathcal P=\{0.01,\dots,0.99\}$. All scores are averaged over the test set. \todo{name the implementation used, e.g.\ \texttt{properscoring}/\texttt{scoringRules}/\texttt{uncertainty-toolbox}, for reproducibility.}

%%%%%%%%%%%%%%%%%%%%%%%%%%%%%%%%%%%%%%%%%%%%%%%%%%%%%%%%%%%%

\end{document}